\documentclass[runningheads]{llncs}

\usepackage{eccv}

\usepackage{eccvabbrv}

\usepackage{graphicx}
\usepackage{booktabs}

\usepackage[accsupp]{axessibility}
\usepackage{multirow}

\usepackage{hyperref}

\usepackage{orcidlink}

\begin{document}

\title{Cross-Domain Generalization in Machine Unlearning via Label-Conditioned Energy Magnitude Regularization} 

\titlerunning{Cross-Domain Generalization in Machine Unlearning}

\author{Syed Ali Ahmed\inst{1}\orcidlink{0009-0009-1736-8980} \and
Syed Bilal Ahsan\inst{1}\orcidlink{0009-0001-5378-8343} \and
Muhammad Zaigham Zaheer\inst{2}\orcidlink{0000-0001-8272-1351}}

\authorrunning{S.~A.~Ahmed et al.}

\institute{National University of Computer and Emerging Sciences, Karachi, Pakistan\\ \email{sayyidaliahmed1@gmail.com}, \email{syed.bilal.ahsan1@gmail.com} \and
Mohamed bin Zayed University of Artificial Intelligence, Abu Dhabi, UAE
\email{zaigham.zaheer@mbzuai.ac.ae}}

\maketitle

\begin{abstract}
Machine unlearning removes the influence of specific data from a trained model. However, most methods treat the forgotten concept as isolated. In this paper, we study what happens to the rest of the model when a class is forgotten, using a label-conditioned energy-based model (EBM) that assigns per-class energies, making the effect directly observable. We forget a class by raising the energy of its image-label pairs, training with a forget term, a retain anchor to the pretrained model, a global margin, and an energy regularizer that stops the energy magnitudes from growing without limit. A propagation term applies the same forget signal to retain samples, weighted by each sample's DINOv2 similarity to the forget class, so forgetting reaches images that resemble it and leaves the rest untouched. We evaluate on two benchmark datasets: 1) On a subset of DomainNet across four visual domains, we forget tiger, lion, and scissors one at a time. Forgetting a class in the sketch domain also erases it from real, clipart, and painting, with forgetting error reaching \textbf{98\%} and \textbf{99\%} for lion and scissors, and the effect carrying over to the most similar class. 2) On CIFAR-10, we turn off the propagation term and forget each of the ten classes on its own. Forgetting is complete (\textbf{100\%}), while the other nine classes retain \textbf{98.5\%} of their pre-unlearning accuracy on average.

\end{abstract}
\keywords{Machine unlearning \and Energy-based models \and Cross-domain generalization}

\section{Introduction}
\label{sec:intro}

Modern classifiers are pretrained on large datasets comprising hundreds if not thousands of classes; however, in practice, these models are rarely used to their complete extent, as most tasks only require a subset of the classifier's knowledge. In addition, these pretrained classifiers may also need to forget specific information that they have learned, due to a privacy request, a safety concern, or a correction to their training data. This process of completely forgetting a piece of information and its effects from a learned model is formally known as `machine unlearning'~\cite{cao2015towards}.

Unlearning methods commonly perform isolated unlearning by identifying a target class, suppressing it, and preserving performance elsewhere~\cite{chen2023boundary, tarun2023fast, chundawat2023can, fan2024salun}. Such formulations typically treat the remaining data as a retain set, without explicitly accounting for semantic relationships between the target and particular retained classes or domains. In practice, however, the information being forgotten is rarely isolated. For instance,  if a model is asked to forget 'tiger' as it appears in hand-drawn sketches, the same object still exists in photographs, paintings, and clipart; forgetting the concept incompletely leaves it fully recoverable through these other domains. Moreover, 'tiger' (regardless of the domain in which it occurs) is not an independent concept. It shares visual and semantic similarity with related classes such as 'lion'. Whether forgetting propagates to such related concepts, and how far it reaches,
is a question that has gathered little attention.

Cheng \textit{et al}. \cite{cheng2026machineunlearningretainforgetentanglement} establish that forgetting a specific class affects semantically related classes, but they treat this as a negative effect that needs to be actively suppressed to preserve retain-set performance. However, depending upon the scenario, suppression may not always be the right choice. In some applications, such as forgetting a person's face for privacy reasons, some spread of forgetting to closely related identities may be preferable to prevent the model from indirectly reconstructing the forgotten identity through highly similar facial features \cite{choi2023towards}. Therefore, the deliberate and controlled propagation of forgetting signals across related domains or classes is an important yet largely overlooked aspect of machine unlearning.

In this paper, we study the propagation of forgetting using label-conditioned Energy Based Models (EBM) instead of a standard softmax classifier. Unlike a softmax classifier, an EBM assigns an explicit energy score to each class for a given input, and the lowest-energy class is the prediction. This lets us reshape the energy landscape around a forgotten class and observe how the energies of other classes and domains shift in response. We show that forgetting a class in one domain generalizes to that class in every other domain: erasing sketch-tigers removes the tiger concept from real, clipart, and painting as well, with no domain-specific retraining. Forgetting also spills to visually similar classes, most strongly the nearest neighbour. This propagation is not solely an effect of our objective. It emerges from the shared representation in any unlearning process.  We therefore focus on controlling how far this spread reaches, not only on showing that it occurs. We take advantage of it with a mechanism based on DINOv2~\cite{oquab2024dinov2} feature similarity and PCA subspace~\cite{jolliffe2002pca} weighting that directs and scales the spread, so that the target class is forgotten across domains and the classes closest to it in feature space receive the strongest forgetting signal.
The main contributions of this paper are summarized as follows:
\begin{itemize}
    \item We introduce a label-conditioned energy-based model (EBM) with energy magnitude regularization to study class forgetting, giving each class an explicit, directly observable energy score.
    \item \textbf{(Finding)} We show that forgetting is not isolated, i.e., erasing a class in one domain removes it from all other domains as well, consistently across forget targets, and the effect further spills to the nearest visually similar classes.
    \item \textbf{(Method)} We introduce a DINOv2 similarity and PCA subspace weighting mechanism, together with a mask, that turns this incidental spread into a controllable one, scaling how far forgetting reaches and steering it toward chosen domains or similar classes.
\end{itemize}

\section{Related Work}
\subsection{Machine Unlearning.}
Machine unlearning aims to remove the influence of specific information from a trained model, ideally producing a model identical to one retrained from scratch on the retained data \cite{shaik2024exploring, xu2024machine, sekhari2021remember}. This field is broadly categorized into two foundational unlearning approaches,
exact unlearning and approximate unlearning. Exact unlearning aims to ensure that the behavior of the updated model matches that of such a retrained model by producing the identical distributions~\cite{xu2023machineunlearningsurvey}. Early work by Cao and Yang~\cite{cao2015towards} described this goal in terms of recomputing model statistics as if the removed data had never been used. Later, Bourtoule \textit{et al}.~\cite{bourtoule2021machine} proposed SISA, which employs a sharding technique that retrains only the subsets of data which contained forget samples, avoiding the need to retrain the entire model. Ginart \textit{et al}.~\cite{ginart2019making} introduced Q-$k$-means, which quantizes cluster centroids so forgetting can be done via metadata updates instead of recomputing the clustering. However, exact unlearning methods are generally based on linear regression, SVMs or leverage $k$-means clustering and do not scale to overparameterized neural nets. This limitation motivates the second broad category, approximate unlearning.

Approximate unlearning trades perfect forgetting for lower computational cost, achieving this via efficient parameter updates through gradient ascent on forget samples~\cite{jang2023knowledge, kurmanji2023towards}. Golatkar \textit{et al}.~\cite{golatkar2020eternal} instead edit the weights directly, they 'scrub' model parameters so that no test performed on the network can distinguish it from one that never saw the forgotten data, without requiring retraining the original training set.

\subsection{Unlearning Across Domains.}
A few recent works consider unlearning in settings where the same class appears
across multiple domains but they approach it from different angles.  Kawamura \textit{et al.} \cite{kawamura2026approximate} propose Approximate Domain Unlearning (ADU), for domain-level forgetting in vision-language models. Instead of removing specific classes, they target an entire domain and suppress its recognition across all classes, effectively removing domain-specific information from the model. Mishra \textit{et al.} \cite{mishra2025selective} keep the class as the unit of forgetting but wrap it around domains. Their training and data-free method removes a target class only from designated domains through a closed-form nullspace projection, while leaving that same class intact in the domains that are retained. Basak and Yin \cite{basak2024forget} use forgetting for a different purpose. They unlearn domain-specific features so that the remaining domain-agnostic features transfer better across domains. Sepahvand \textit{et al.} \cite{sepahvand2025selective} treat the forget set and a held-out validation set as two 'domains' and apply gradient reversal to erase forget-set representations. In their case domain refers to distinct data distributions, not to visual domains such as sketch or photograph, and the method does not deal with forgetting across styles. None of these works, however, examine whether forgetting a class in one domain spreads to that same class in other domains or other semantically related classes. This is the question we study.

\subsection{Energy-Based Models.}
Energy-based models (EBMs) assign a scalar energy to each input and define an unnormalized distribution in which likely samples take low energy \cite{lecun2006tutorial}. They have been
applied to image generation~\cite{xie2016theory, nijkamp2020anatomy, du2020compositional, gao2021learning}, structured
prediction~\cite{belanger2016structured, tu2019benchmarking}, and out-of-distribution
detection~\cite{liu2020energy, wang2021energy}.
Grathwohl \textit{et al.} \cite{duvenaud2020your} showed that a standard classifier can be employed as an EBM, since its logits can be reinterpreted as a joint energy over inputs and labels (JEM); this energy is most often used generatively, producing samples through Langevin dynamics \cite{du2019implicit}. Li \textit{et al.}~\cite{li2022energy} utilize EBMs for continual learning. They perform selective incremental updates by lowering the energy
of an input at its true label and raising it at selected negatives, so that updates touch only the chosen classes and forgetting is avoided. Recent unlearning work builds directly on this view: DSDA \cite{wang2025efficient} reinterprets a classifier's logits as an energy function and runs Langevin sampling to synthesize surrogate training data, enabling source-free unlearning without the original dataset.



\section{Method}
In this paper, we propose reshaping the energy landscape of a pretrained, label-conditioned energy-based model to capture the forgetting of a class within a specific domain. The motivation builds on the standard EBM formulation proposed by LeCun \textit{et al}.~\cite{lecun2006tutorial}, which scores every $(X, Y)$ pair with a single scalar value called energy where $X$ is an input image and $Y$ is the label associated with it. The idea is that a correct $(X, Y)$ pair should have lower energy compared to an incorrect or corrupted $(X, Y)$ pair. We adopt EBMs instead of a softmax classifier because forgetting, in our setting, means making specific image-label pairs unfavorable, and an energy gives one quantity per pair that we can push up to forget. This gives us one direct mechanism and one direct measurement built on the same quantity: we raise the energy of the forget cell while anchoring the rest of the model to its original behavior, and we later read the resulting energies to visualise forgetting on CIFAR-10, as shown in Figure~\ref{fig:energy}.

\begin{figure*}[t]
    \centering
    \includegraphics[
        width=\textwidth,
        trim=50 35 50 35,
        clip
    ]{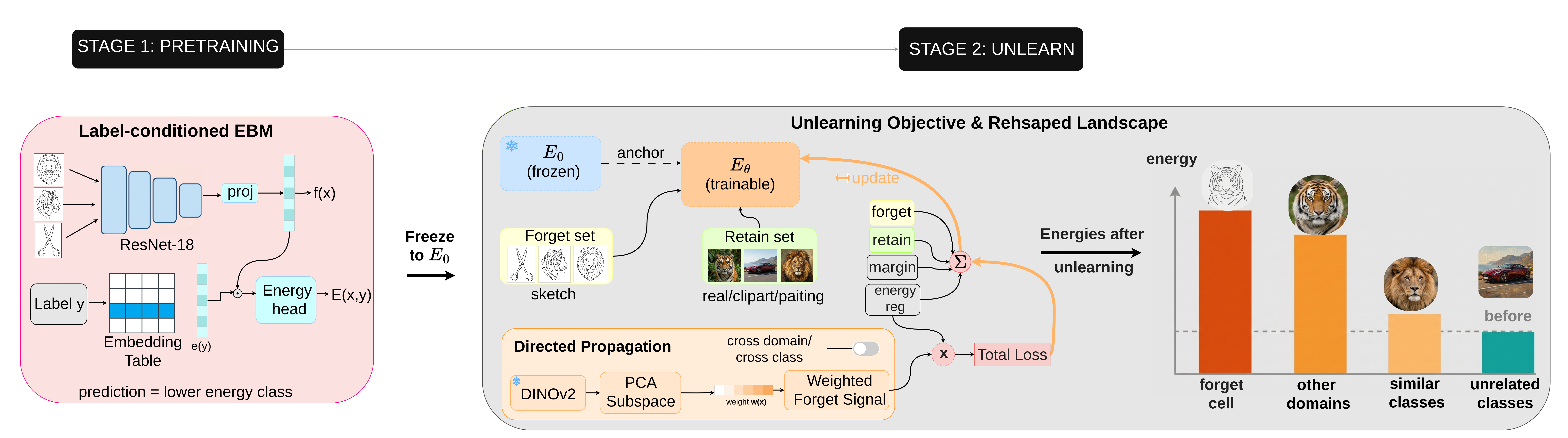}
    \caption{Overview of our method. Stage~1 pretrains a label-conditioned EBM to obtain frozen reference $E_0$. Stage~2 trains a copy $E_\theta$ to forget a single (class, domain) cell by combining four losses (forget, retain, margin, and energy regularization) with a DINOv2 similarity-weighted propagation mechanism. Scope masking controls whether the forget signal propagates across domains or semantically related classes, selectively forgetting target knowledge while preserving unrelated concepts.}
    \label{fig:main}
\end{figure*}

The pipeline has two stages, shown in Figure~\ref{fig:main}. We first pretrain the EBM on all classes and domains with a contrastive energy loss, producing a reference model $E_0$ (Section~\ref{sec:arch}). We then freeze $E_0$ and train a copy $E_\theta$ to forget. The unlearning objective increases the energy of samples belonging to the forget set while anchoring retain energies to $E_0$. To enable this forgetting to propagate to other domains of the target class and to other similar classes, we add a directed propagation term: each retain sample is weighted by how strongly its DINOv2 features align with the forget cell's feature subspace, and a similarity-weighted forget signal is applied in proportion to that weight, though some propagation occurs even without this added term, since it also arises from the shared backbone alone (Section~\ref{sec:propagation}). A masking strategy is used alongside this to control the reach of the signal, either confining it to the forget class in other domains or letting it act across classes. The result is a model that no longer recognizes the forget set, with forgetting also propagating to related classes and domains, as we study in the subsequent sections.

\subsection{Problem Formulation}

Each image $x$ is associated with a class label $c \in \{1,\dots,K\}$ and a domain label
$d \in \{1,\dots,M\}$. In our experiments, we set $K=26$ and $M=4$. We randomly sample 26
of the 345 DomainNet classes and use four of its six domains, real, sketch, clipart, and
painting. The energy model $E_\theta$ assigns a scalar energy to an image $x$ paired with a
\emph{candidate} label $y \in \{1,\dots,K\}$, and is conditioned on the class label only;
here $y$ is the query label whose energy we evaluate, not necessarily the true class $c$.
The domain label is never given to the network and is used solely to define what we forget.
A class is predicted by taking the candidate label of lowest energy,
\begin{equation}
\label{eq:argmin}
\hat{y} = \arg\min_{y \in \{1,\dots,K\}} E_\theta(x, y),
\end{equation}
so the prediction $\hat{y}$ is correct when $\hat{y}=c$.

We forget a single $(\text{class}, \text{domain})$ cell rather than a whole class or a whole domain. For a forget class $c^\star$ and forget domain $d^\star$, the forget set is
\begin{equation}
D_f = \left\{(x, c^\star) \;\middle|\; \mathrm{class}(x)=c^\star,\ \mathrm{domain}(x)=d^\star \right\},
\end{equation}
and the retain set $D_r$ is its complement, i.e., every remaining image-class pair in the collection. Taking $c^\star=\text{tiger}$ and $d^\star=\text{sketch}$, the goal is to stop the model recognizing sketch tigers while its behavior on tigers in real, clipart, painting, and on every non-tiger class, stays as it was. The forget class is not removed from the retain set entirely, i.e., images of tiger from the other domains remain in $D_r$. This overlap is intentional. It is what makes cross-domain forgetting measurable, and the propagation mechanism in Section~\ref{sec:propagation} operates particularly on these retain samples.

\subsection{Base architecture and pretraining}
\label{sec:arch}

The energy model turns an (image~$x$, label~$y$) pair into a single scalar score by combining an image feature with the label's embedding. We use a ResNet-18 backbone to encode the image, followed by a linear projection, $f(x) = \mathrm{proj}(\mathrm{ResNet18}(x)) \in \mathbb{R}^{d_e}$,
while each class label is mapped to a vector $e(y) \in \mathbb{R}^{d_e}$ through a learned
embedding table. The two are multiplied elementwise and passed through a linear head to produce the scalar energy,
\begin{equation}
\label{eq:energy}	
E_\theta(x, y) = w^\top \left(f(x) \odot e(y)\right) + b,
\end{equation}
where $w \in \mathbb{R}^{d_e}$ and $b \in \mathbb{R}$ are the learned weight vector and bias
of the linear energy head, and $d_e = 128$ in all experiments. The projection maps the image into a shared $d_e$-dimensional embedding space so that it can be combined
with the label embedding $e(y)$, which lives in the same space. This score is unnormalized and specific to a single (image, label) pair, which is the quantity the unlearning objective later raises for the forget cell and holds fixed for the rest.

The backbone is initialized from ImageNet-pretrained weights, and we do not train all of it. During pretraining the last residual stage (layer4) is fine-tuned along with the projection, the label embedding, and the energy head, while the earlier layers are kept frozen. This keeps low-level features fixed and adapts only the higher-level representation to DomainNet. The number of trainable stages is a single setting we vary in later experiments, and a low-rank (LoRA) variant that freezes the convolutional weights and trains only rank-$r$ adapters is available as an alternative.

We pretrain $E_\theta$ on all 26 classes across the four domains with a supervised energy-contrast loss. For each image, the loss compares the energy of its correct label against $N$ sampled incorrect labels and penalizes any case where the correct label is not lower by a margin $m_0$. Taking the hardest of the $N$ negatives per image, the loss is
\begin{equation}
\label{eq:pretrain}	
\mathcal{L}_{\text{pre}} = \frac{1}{|B|} \sum_{x \in B} \max_{j=1,\dots,N} \mathrm{ReLU}\left(m_0 + E_\theta(x, c) - E_\theta(x, c_j^-)\right),
\end{equation}
where $c$ is the true class, $c_j^- \neq c$ are the sampled negatives, and $B$ is the minibatch. We use $N=10$ and $m_0=1$, optimize with Adam, and stop early on validation accuracy, with the remaining settings given in our implementation details. The model produced by this stage is the reference $E_0$. It is held frozen for the rest of the method.

\subsection{Unlearning Objective}
\label{sec:unlearn}

Unlearning trains a copy $E_\theta$, initialized from the pretrained reference $E_0$, while $E_0$ stays frozen. The objective moves the energy landscape in two directions at once. It raises the energy of the forget cell at its correct label until that label is no longer the lowest, so by Eq~\eqref{eq:argmin} the model settles on some different classes for those images and no longer recognizes them.

We design four loss functions for this task and every training step applies all of them together. The forget term raises the energy of each forget image at its correct label above the energies it assigns to the other labels. The retain term ties the rest of the model back to $E_0$. A margin term enforces a consistent gap between forget and retain energies, and a small regularizer keeps the energy magnitudes bounded so the other terms cannot inflate them without limit. At each step we draw a forget minibatch from $D_f$ and a retain minibatch from $D_r$, evaluate the four terms on them, and update $E_\theta$ by their weighted sum while $E_0$ is held fixed. We describe each term below and then give the combined objective.

\subsubsection{Forget term.}
The forget term acts on images from $D_f$, each associated with the forget class $c^\star$. For each image, it requires the energy of the correct pair to exceed that of every other label by a margin $m$. The maximum selects the label for which this is hardest to satisfy, i.e., the label with the highest energy, and meeting that one constraint implies all the others are met as well:
\begin{equation}
\label{eq:forget}	
    L_f = \frac{1}{|B_f|}\sum_{x\in B_f}\ \max_{y'\neq c^\star}\ \mathrm{softplus}\left(m + E_\theta(x, y') - E_\theta(x, c^\star)\right).
\end{equation}
This mirrors the pretraining loss of Eq~\eqref{eq:pretrain} with the sign reversed. Pretraining picks the lowest-energy negative to push the correct label to the bottom of the ranking, whereas the forget term picks the highest-energy competitor and drives the forget cell's correct label to the top, so by Eq~\eqref{eq:argmin} the correct label becomes the label the model is least likely to output.

\subsubsection{Retain anchor.}
The retain term is where $E_0$ acts as an anchor. For each retain image it keeps the current energy close to the value $E_0$ assigned, by penalizing the squared difference between the two. We divide by the average pretrained magnitude so the penalty does not depend on how large the energies have grown:
\begin{equation}
\label{eq:retain}	
    L_r = \left.\frac{1}{|B_r|}\sum_{x\in B_r}\big(E_\theta(x, c) - E_0(x, c)\big)^2 \middle/ \frac{1}{|B_r|}\sum_{x\in B_r}\big|E_0(x, c)\big|\right.,
\end{equation}
where $c$ is the true class of $x$. Since $D_r$ holds the forget class in the other domains, this term is also what keeps the model's behavior on the forget class (`tiger' in our example) in real, clipart, and painting close to its pretrained state, until the mechanism of Section~\ref{sec:propagation} deliberately relaxes it.

\subsubsection{Global margin.}
The margin term adds a global separation between the two sets. It pushes forget energies to lie above retain energies by the same margin $m$, taken over a forget and a retain minibatch of equal size:
\begin{equation}
\label{eq:margin}	
    L_m = \frac{1}{|B_f|}\sum_{i=1}^{|B_f|} \mathrm{softplus}\left(m + E_\theta(x_r^{(i)}, c_r^{(i)}) - E_\theta(x_f^{(i)}, c_f^{(i)})\right),
\end{equation}
where the $i$-th retain example $(x_r^{(i)}, c_r^{(i)})$ and forget example $(x_f^{(i)}, c_f^{(i)})$ are paired across the two equal-size minibatches ($|B_f| = |B_r|$).

\subsubsection{Energy Magnitude Regularization.}
The final regularization term penalizes large energy magnitudes, discouraging the forget and margin objectives from driving the energies to arbitrarily large values:

\begin{equation}
\label{eq:ereg}	
    L_e = \tfrac{1}{2}\left(\frac{1}{|B_f|}\sum_{x_f\in B_f} E_\theta(x_f, c_f)^2 + \frac{1}{|B_r|}\sum_{x_r\in B_r} E_\theta(x_r, c_r)^2\right).
\end{equation}

\subsubsection{Full objective.}
The four terms are combined by fixed weights,
\begin{equation}
\label{eq:obj}	
    L = \lambda_f L_f + \lambda_r L_r + \lambda_m L_m + \lambda_e L_e,
\end{equation}
where the weights $\lambda_f, \lambda_r, \lambda_m, \lambda_e$ balance the four terms, and the same margin $m$ is shared by the forget, margin, and propagation terms. We report their values in the implementation details. Section~\ref{sec:propagation} adds one further term that governs how far this forgetting propagates across domains.

\subsection{Directed Propagation}
\label{sec:propagation}

We also design a mechanism that directs and amplifies how far forgetting spreads beyond the forget cell, i.e., how forgetting carries across domains. It applies a forgetting signal to retain samples, weighted by their resemblance to the forget cell, so samples that resemble the forget cell are pushed hardest while those with little in common are left almost untouched. A mask then fixes the scope of this push, either restricting it to the forget class in other domains or opening it to other classes. A part of this spread also occurs on its own, since the backbone and label embedding are shared.

\subsubsection{Similarity weighting.}
We use the frozen DINOv2 image encoder to measure the similarity between each retain sample and the forget cell. Let $\phi(x)$ be the $L_2$-normalized DINOv2 feature of image $x$ and $\mu$ the mean feature over the forget and retain sets. We then apply PCA to the centered forget-cell features and stack the top $k$ principal directions as the rows of $S$. Each retain sample is then weighted by the fraction of its centered feature that lies in this forget subspace:
\begin{equation}
\label{eq:weight}	
    \alpha(x) = \mathrm{clip}\!\left(\frac{\lVert S\,(\phi(x) - \mu)\rVert^2}{\lVert \phi(x) - \mu\rVert^2},\ 0,\ 1\right),
\end{equation}
with $\mathrm{clip}(z,0,1) = \min(\max(z,0),1)$. A weight near one means the sample looks much like the forget cell, and a weight near zero means it does not. We compute these weights once, before unlearning, and keep them fixed.

\subsubsection{Weighted forget signal.}
This term applies the same forget-style push from Eq~\eqref{eq:forget}, but applies it to retain samples and scales it by the weight $\alpha(x)$. For a retain sample with true class $c$, we pick one random incorrect label $c^-$ and push the correct-label energy $E_\theta(x, c)$ above $E_\theta(x, c^-)$ by the margin $m$:
\begin{equation}
\label{eq:prop}	
    L_p = \frac{1}{|B_r|}\sum_{x\in B_r} \alpha(x)\, \mathrm{relu}\left(m + E_\theta(x, c^-) - E_\theta(x, c)\right).
\end{equation}
The difference from Eq~\eqref{eq:forget}     is small: we use a single sampled negative and a plain hinge (relu) instead of a softplus over all labels. What matters is the weight $\alpha(x)$. A sample close to the forget cell ($\alpha$ near 1) gets a strong push, so the model begins to forget it the way it forgets the cell. A sample far from it ($\alpha$ near 0) gets almost no push, and the retain anchor of Eq~\eqref{eq:retain} keeps its energy at the pretrained value.

\subsubsection{Reach of the signal.}
A mask on the weights decides which samples are eligible. To allow unlearning to be propagated across other domains of the target class, we keep $\alpha(x)$ only for retain samples whose class is the forget class and zero the rest, so the signal reaches tigers in real, clipart and painting but nothing else, and forgetting stays limited to the target class across domains. In the cross-class setting we keep the weights for all retain samples, so the signal also reaches other classes in proportion to their similarity and forgetting is allowed to spread to visually related categories. The two settings use the same weighting and differ only in which weights are set to zero.
The propagation term is added to the objective of Section~\ref{sec:unlearn},
\begin{equation}
\label{eq:total}	
    L_{\text{total}} = L + \lambda_p L_p.
\end{equation}
Setting $\lambda_p = 0$ does not switch propagation off because part of the spread comes from the shared backbone and not this term alone, its extent also depends on how much of the backbone adapts, set by the trainable stages or the LoRA rank.

\begin{table}[t]\centering
\caption{Unlearning on DomainNet across three forget targets, each forgotten in the sketch domain. F is forgetting error ($100~\text{acc}$, \%, $\uparrow$) on the target cell (F/sk.) and on the same class in the other three domains (F/oth.). Mem is retain accuracy before and after unlearning ($\uparrow$).}
\label{tab:main}
\begin{tabular}{@{}llcccc@{}}
\toprule
Target & Method & F/sk.\,$\uparrow$ & F/oth.\,$\uparrow$ & Mem$_{\text{pre}}$ & Mem$_{\text{unl}}$\,$\uparrow$ \\
\midrule
\multirow{4}{*}{tiger}
 & Retrain  & 19.9 & 3.3   & 96.0 & 96.2 \\
 & NegGrad  & 100.0& 100.0 & 96.0 & 85.3 \\
 & Finetune & 12.2 & 2.6   & 96.0 & 97.2   \\
 & Ours & 96.1 & 83.7 & 96.0 & 76.7  \\
\midrule
\multirow{4}{*}{lion}
 & Retrain  & 46.7 & 4.1   & 95.9 & 96.1  \\
 & NegGrad  & 100.0& 100.0 & 95.9 & 70.0  \\
 & Finetune & 18.2 & 1.9   & 95.9 & 97.5 \\
 & Ours & 99.4 & 98.5 & 95.9 & 80.9  \\
\midrule
\multirow{4}{*}{scissors}
 & Retrain  & 54.9 & 4.5   & 96.1 & 97.2  \\
 & NegGrad  & 100.0& 100.0 & 96.1 & 72.4  \\
 & Finetune & 36.6 & 1.4   & 96.1 & 96.3  \\
 & Ours & 100.0 & 99.2 & 96.1 & 77.2 \\
\bottomrule
\end{tabular}
\end{table}

\begin{table}[t]
\centering
\caption{Isolated unlearning on CIFAR-10, all ten classes as forget targets. F is forgetting error ($100-\text{acc}$, \%). Mem is reported before (pretrained) and after unlearning on the other nine classes; the pretrained model tops out near 70\% on this backbone, so the near-unchanged Mem column reflects preserved utility, not degradation. Utility is the after/before ratio; retain energy preservation is the rank-order agreement of retain energies with the pretrained model ($1.0$ = identical); MIA is membership-inference AUC on forget samples ($0.5$ = indistinguishable).}
\label{tab:cifar10}
\begin{tabular}{@{}lcccccc@{}}
\toprule
Class & F\,$\uparrow$ & Mem$_{\text{pre}}$ & Mem$_{\text{unl}}$\,$\uparrow$ & Utility\,$\uparrow$ & Energy pres.\,$\uparrow$ & MIA\,($\to$0.5) \\
\midrule
airplane   & 100 & 69.3 & 67.8 & 97.9  & 0.862 & 0.489 \\
automobile & 100 & 69.3 & 66.4 & 95.7  & 0.872 & 0.496 \\
bird       & 100 & 70.1 & 71.8 & 102.5 & 0.885 & 0.491 \\
cat        & 100 & 71.0 & 72.5 & 102.1 & 0.864 & 0.490 \\
deer       & 100 & 70.9 & 70.0 & 98.8  & 0.857 & 0.508 \\
dog        & 100 & 70.6 & 70.4 & 99.8  & 0.813 & 0.484 \\
frog       & 100 & 69.2 & 69.1 & 99.9  & 0.840 & 0.508 \\
horse      & 100 & 69.6 & 67.8 & 97.4  & 0.839 & 0.498 \\
ship       & 100 & 69.0 & 66.0 & 95.6  & 0.852 & 0.505 \\
truck      & 100 & 68.7 & 65.6 & 95.5  & 0.853 & 0.491 \\
\midrule
mean       & 100 & 69.8 & 68.7 & 98.5  & 0.854 & 0.496 \\
\bottomrule
\end{tabular}
\end{table}

\begin{figure*}[t]
\centering
\includegraphics[width=0.9\textwidth]{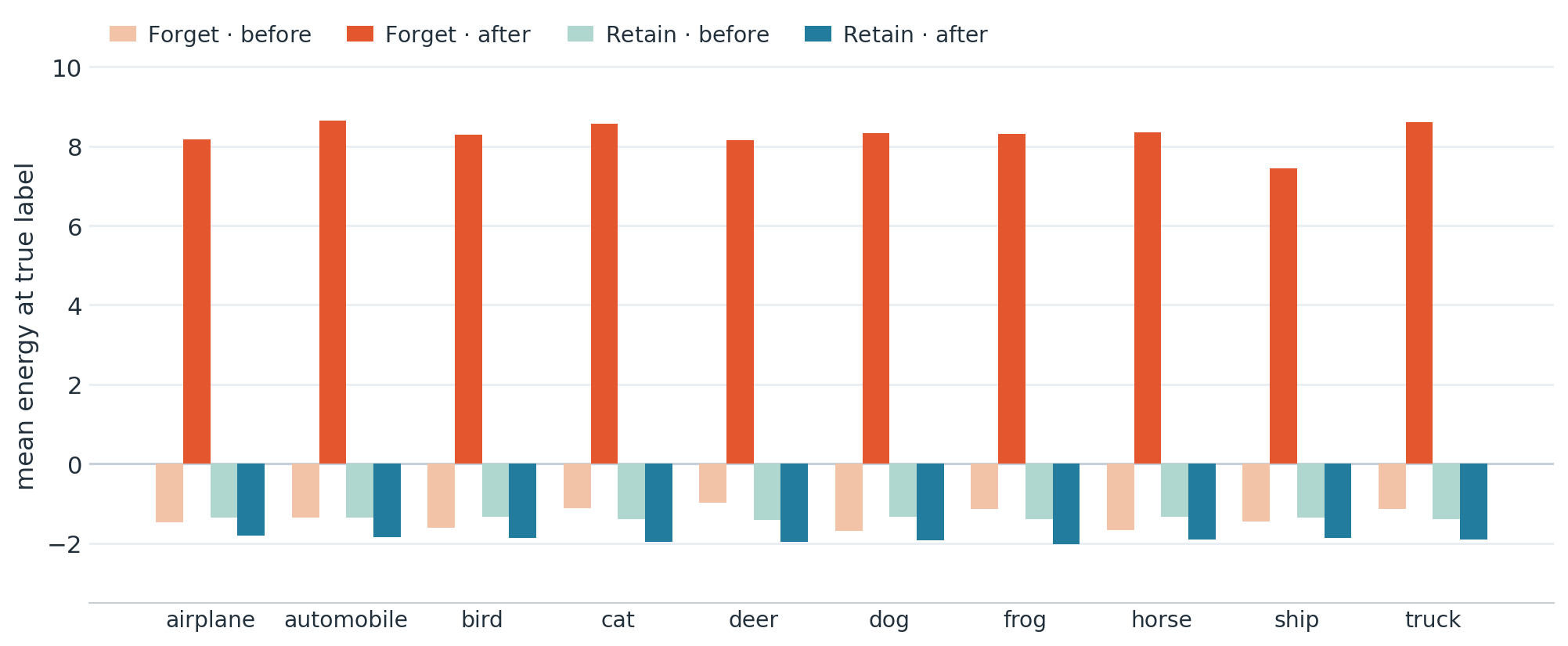}
\caption{Energy-landscape reshaping under isolated unlearning across all ten CIFAR-10 classes. Bars show the mean energy at the true label for the forget class and the retain set, before and after unlearning. For every class the forget-class energy is lifted by roughly ten units (to $\approx\!+8.3$), while the retain-set energy is essentially unchanged ($\approx\!-1.4\to-1.9$). This is the mechanism behind the energy gaps: forgetting raises only the target's energy and leaves the retain landscape intact.}
\label{fig:energy}
\end{figure*}

\begin{table}[t]\centering
\caption{Cross-class spillover (DomainNet). When a class is forgotten, degradation concentrates on its nearest visual neighbour. NN is the retain class with highest DINOv2 similarity to the forget class; ``NN drop'' is its accuracy drop after unlearning. $r$ and $\rho$ are the Pearson and Spearman correlation between a retain class's similarity to the forget class and its accuracy drop, over all 25 retain classes.}
\label{tab:spillover}
\begin{tabular}{@{}lcccc@{}}
\toprule
Forget target & nearest neighbour (sim.) & NN drop & $r$ & $\rho$ \\
\midrule
tiger    & lion (0.55)   & $-75\%$ & 0.76 & 0.44 \\
scissors & pliers (0.78) & $-46\%$ & 0.42 & 0.10 \\
lion     & tiger (0.56)  & $-9\%$  & 0.10 & 0.21 \\
\bottomrule
\end{tabular}
\end{table}

\section{Results}

\subsection{Experimental Setup}

\noindent\textbf{Datasets.} We use a 26-class subset of DomainNet~\cite{peng2019moment} across four domains: real, sketch, clipart, and painting. We forget three targets in turn: tiger, lion, and scissors, each in the sketch domain. The 25 retain classes span a range of DINOv2 similarity to the forget class, so degradation can be measured from highly similar down to unrelated classes. For isolated single-class forgetting, we additionally evaluate all ten classes of CIFAR-10.

\begin{table}[t]\centering
\caption{What controls propagation (DomainNet, tiger/sketch). Backbone capacity and the propagation term each move the similarity--degradation correlation. As more of the backbone adapts, propagation ($r$) rises monotonically, and with the term off ($\lambda_p{=}0$) forgetting still spreads, confirming it is partly intrinsic to the shared backbone.}
\label{tab:control}
\begin{tabular}{@{}llcccc@{}}
\toprule
Variant & trainable & F\,$\uparrow$ & Mem\,$\uparrow$ & $r$\,$\uparrow$ & $\rho$\,$\uparrow$ \\
\midrule
\multicolumn{6}{@{}l}{\emph{Backbone capacity} ($\lambda_p{=}3$)} \\
Frozen (heads only)   & heads only & 60.6 & 92.1 & 0.52 & 0.22 \\
LoRA r8, layer4       & 157k       & 66.6 & 89.3 & 0.535 & 0.083 \\
LoRA r8, layer3+4     & 227k       & 62.7 & 87.0 & 0.642 & 0.319 \\
Full layer4 (Ours)    & 8.5M       & 96.1 & 77.9 & \textbf{0.76} & \textbf{0.44} \\
\midrule
\multicolumn{6}{@{}l}{\emph{Propagation term} (full capacity)} \\
$\lambda_p{=}0$ (isolated)      & 8.5M & 73.6 & 82.8 & 0.63 & 0.36 \\
$\lambda_p{=}3$ (Ours)          & 8.5M & 96.1 & 77.9 & \textbf{0.76} & \textbf{0.44} \\
\bottomrule
\end{tabular}
\end{table}

\begin{figure}[t]
\centering
\includegraphics[width=0.85\linewidth]{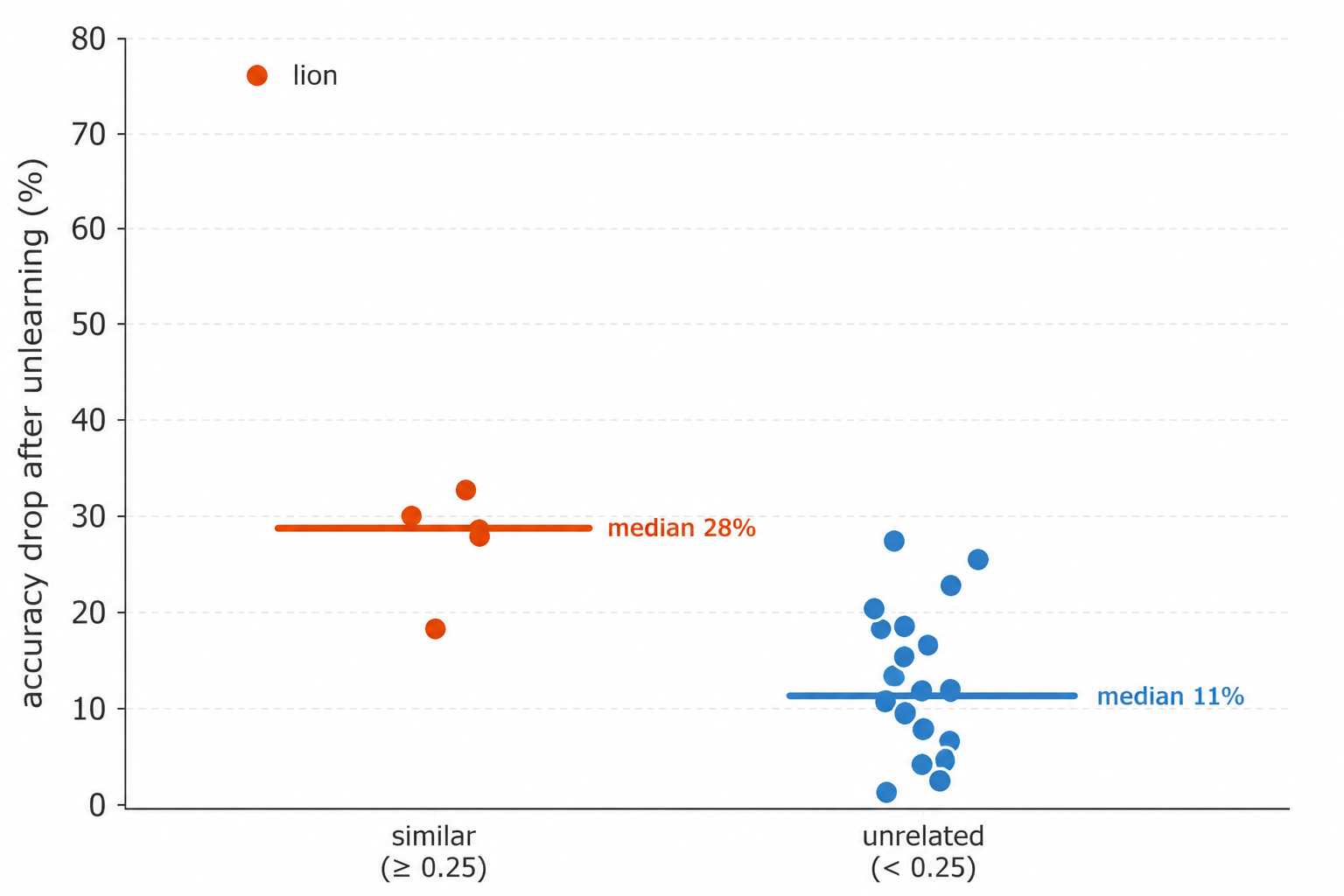}
\caption{Cross-class spillover for tiger/sketch. Each point is a retain class's accuracy drop after unlearning, grouped by DINOv2 similarity to tiger (threshold $0.25$). Classes above the threshold lose a median of $28\%$ against $11\%$ for the rest, and lion, tiger's nearest neighbour, degrades most. Bars mark group medians.}
\label{fig:propagation}
\end{figure}

\noindent\textbf{Model.} The label-conditioned EBM of Sec.~\ref{sec:arch} with a ResNet-18 backbone, used for both datasets.

\noindent\textbf{Baselines.} Retrain (from scratch on the retain set, gold reference), NegGrad (gradient ascent on the forget set), and Finetune (continued training on the retain set). On DomainNet we run our method with the cross-class (propagate) mask; on CIFAR-10 we run it in the isolated single-class setting.

\noindent\textbf{Metrics.} On DomainNet, we report post-unlearning top-1 accuracy. Forgetting is split into the target cell (tiger/sketch) and the same class in other domains. Retain-set accuracy, which we report as Mem, is measured over all retain classes before and after unlearning. The spread to other classes is measured as the Pearson $r$ (and Spearman $\rho$) between a retain class's DINOv2 similarity to the forget class and its accuracy drop, reported in Table~\ref{tab:spillover}. On CIFAR-10, we report the forgetting rate, model utility (Mem after unlearning divided by Mem before), retain energy preservation (rank agreement of retain energies with $E_0$), and the membership-inference AUC on forget samples, where $0.5$ indicates the forgotten samples are indistinguishable from data the model never saw.

\noindent\textbf{Implementation.} Pretraining: 50 epochs, Adam, lr $10^{-4}$, $N=10$, $m_0=1$, early stopping on validation accuracy. DomainNet unlearning: 1500 steps, batch 32, lr $10^{-4}$, weights $\lambda_f=1,\lambda_r=15,\lambda_m=1,\lambda_e=10^{-3}$, margin $m=5$; propagation $\lambda_p=3$, $k=5$, frozen DINOv2 ViT-B/14; 20\% holdout for MIA. CIFAR-10 unlearning uses the same four-term objective without the propagation term ($\lambda_p=0$).

\subsection{Forgetting Generalizes Across Domains}
Table~\ref{tab:main} shows that our method drives F/oth.\ close to zero for lion and scissors and to ${\sim}16\%$ for tiger (98.5, 99.2, 83.7), meaning removing sketch tigers also removes them from real, clipart, and painting. Retrain and Finetune barely reduce the class in the other domains (F/oth.\ $<5$). Since the class still exists there, the model keeps recognizing it in sketch as well, so retraining on the retain set simply relearns it from the other domains and the deletion never takes place. NegGrad also forgets across domains, suggesting this effect comes from the shared backbone rather than our loss, but it is less selective and reduces retain accuracy (70--72 on lion and scissors) and it has no way to limit how far forgetting reaches. Our method keeps more of this accuracy while still forgetting across domains, at a moderate cost.

\subsection{Isolated Forgetting on CIFAR-10}

On DomainNet, forgetting is allowed to spread across classes. Isolated forgetting is the opposite case where we remove a single class and leave the others unchanged despite them being highly similar to the target class/domain. We evaluate it on CIFAR-10 by applying the same unlearning procedure to each of the ten classes in turn.

Table~\ref{tab:cifar10} reports all ten runs. We achieve 100\% forget error on all classes, while Mem stays close to the pretrained baseline of about 70\% on CIFAR-10, within about one point on average. This gives a mean model utility of 98.5\% and a minimum of 95.5\%. The membership-inference AUC averages 0.496, indistinguishable from the chance value of 0.5, so forget-set membership cannot be recovered from the unlearned model.

Figure~\ref{fig:energy} reports the same ten runs in terms of energy. Before unlearning, the forget class and the retain set both have mean energies closer to mean energy $-1.4$. After unlearning, the forget-class energy rises to $+8.3$ while the retain energy barely moves, from $-1.4$ to $-1.9$. Because inference is $\arg\min_y E_\theta(x,y)$, raising only the forget-class energy above the retain energies removes the forget class from the prediction and leaves the other decisions unchanged. The retain energy preservation in Table~\ref{tab:cifar10} is ($\approx 0.85$). This means the retain classes keep almost the same energy ordering after unlearning as before.

\subsection{Propagation to Similar Classes}

Forgetting spreads to other classes too, but the effect is small and affects mostly the nearest neighbour. The most affected retain class is always the one closest to the target in DINOv2 space. Lion drops 75\% when tiger is forgotten, and pliers drops 46\% when scissors is forgotten. Over all 25 retain classes the drop still correlates with similarity, with $r=0.76$ for tiger, but most of that correlation comes from the single nearest class and the strength is not the same across targets. Figure~\ref{fig:propagation} shows this for tiger, where classes above the similarity threshold lose a median of 28\% and the rest lose 11\%. Section~\ref{sec:control} shows how the weighting term controls how far this spread reaches.

\subsection{Controlling Propagation}
\label{sec:control}

We show in Table~\ref{tab:control} that two factors can change how far forgetting spreads. The first is how much of the backbone we train. When the backbone is frozen and only the heads are updated, the similarity--degradation correlation is low ($r=0.52$). Training small LoRA adapters raises it, and training the full last layer raises it to $0.76$. The forgetting therefore spreads through the shared backbone, and it reaches further when more of the backbone is allowed to adapt. The second is the propagation term. Setting it from $\lambda_p{=}0$ to $\lambda_p{=}3$ raises the correlation from $r=0.63$ to $r=0.76$ and pushes the spread toward the classes most similar to the forget class.

\section{Conclusion}
In this study, we examine how forgetting propagates when a model is made to drop a single class. A label-conditioned EBM made this easy to observe, since every class carries its own energy, letting us track how class-specific energies change rather than only verifying removal of the target. Forgetting is not fixed to where it is applied: removing a class from the sketch domain significantly drops accuracy on that class in real, clipart, and painting, though those samples are not in the forget set, and this held across all three targets. We also observe spillover to visually similar classes, strongest for nearest neighbours, though the magnitude varies across targets. Our approach controls how far this spreads, using DINOv2 similarity to determine where the forgetting signal is applied; disabling this component reduces the objective to isolated unlearning.

\noindent\textbf{Limitations and future work.} The spread to similar classes concentrates on the nearest neighbour and changes size with the target, so it is better read as a control to turn up or down than a precise setting. Removing a class also lowers accuracy slightly on retained classes, expected since shared features mean erasing one disturbs the others. Our experiments use one DomainNet subset and CIFAR-10 on small backbones, so they show the effect is real and steerable, not how far it can ultimately be pushed. Larger backbones, more datasets, and settings where controlled spread is the goal, such as removing a person's identity together with its closest look-alikes, are the natural next steps.

%
%
\bibliographystyle{splncs04}
\bibliography{main}
\end{document}